\documentclass[10pt,twocolumn,letterpaper]{article}

\usepackage{dingbat} 

\usepackage[pagenumbers]{cvpr} % To force page numbers, e.g. for an arXiv version

\usepackage[dvipsnames]{xcolor}

\usepackage{amssymb}
\usepackage{pifont} 
\usepackage{tikz}

\usepackage[pdftex,outline]{contour}
\contourlength{0.8pt}% Thickness of copies
\contournumber{1}% How many copies

\definecolor{cvprblue}{rgb}{0.21,0.49,0.74}
\usepackage[pagebackref,breaklinks,colorlinks,citecolor=cvprblue]{hyperref}

\def\paperID{10819}
\def\confName{CVPR}
\def\confYear{2024}

\title{pix2gestalt: Amodal Segmentation by Synthesizing Wholes}

\author{Ege Ozguroglu$^1$\ \ Ruoshi Liu$^1$\ \ Dídac Surís$^1$\ \ Dian Chen$^2$\ \ Achal Dave$^2$\ \ Pavel Tokmakov$^2$\ \ Carl Vondrick$^1$ 
\vspace{0.2cm}
\\$^1$Columbia University \ \ $^2$Toyota Research Institute
\vspace{0.1cm}
\\
\href{https://gestalt.cs.columbia.edu/}{\textbf{\url{gestalt.cs.columbia.edu}}}
\vspace{-0.32cm}
}

\begin{document}
\maketitle
\begin{abstract}
\vspace{-0.4cm}
We introduce pix2gestalt, a framework for zero-shot amodal segmentation, which learns to estimate the shape and appearance of whole objects that are only partially visible behind occlusions. By capitalizing on large-scale diffusion models and transferring their representations to this task, we learn a conditional diffusion model for reconstructing whole objects in challenging zero-shot cases, including examples that break natural and physical priors, such as art. As training data, we use a synthetically curated dataset containing occluded objects paired with their whole counterparts. Experiments show that our approach outperforms supervised baselines on established benchmarks. Our model can furthermore be used to significantly improve the performance of existing object recognition and 3D reconstruction methods in the presence of occlusions. 
\end{abstract}   
\vspace{-0.5cm}
\section{Introduction}

\label{sec:intro}
\begin{figure*}[t]
\centering \includegraphics[width=0.9652\textwidth]{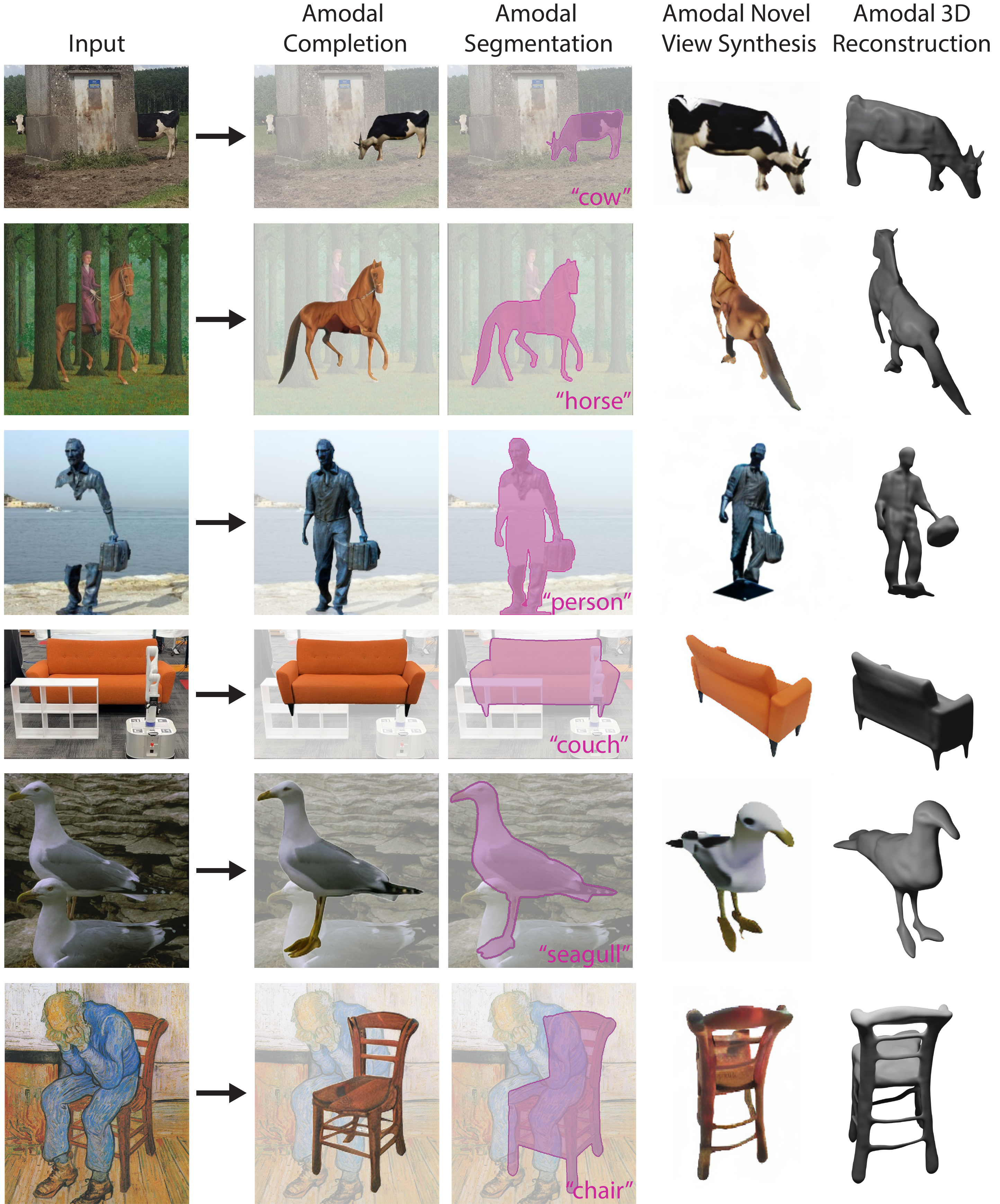}
        \captionof{figure}{\textbf{Amodal Segmentation and Reconstruction via Synthesis.} We present \textit{\textbf{pix2gestalt}}, a method to synthesize whole objects from only partially visible ones, enabling amodal segmentation, recognition, novel-view synthesis, and 3D reconstruction of occluded objects. \vspace{0.9cm}}
        \label{fig:teaser}
\end{figure*}

Although only parts of the objects in Figure~\ref{fig:teaser} are visible, you are able to visualize the whole object, recognize the category, and imagine its 3D geometry. Amodal completion is the task of predicting the whole shape and appearance of objects that are not fully visible, and this ability is crucial for many downstream applications in vision, graphics, and robotics. Learned by children from an early age \cite{piaget2013construction}, the ability can be partly explained by experience, but we seem to be able to generalize to challenging situations that break natural priors and physical constraints with ease. In fact, we can imagine the appearance of objects during occlusions that cannot exist in the physical world, such as the horse in Magritte's \textit{The Blank Signature}.

What makes amodal completion challenging compared to other synthesis tasks is that it requires
grouping for both the visible and hidden parts of an object.
To complete an object, we must be able to first recognize the object from partial observations, then synthesize only the missing regions for the object. 
Computer vision researchers and gestalt psychologists have extensively studied amodal completion in the past \cite{ehsani2018segan,zhan2020self,zhu2016semantic,ke2021deep,qi2019amodal,reddy2022walt, ling2020variational,kar2015amodal}, creating models that explicitly learn figure-ground separation. However, the prior work has been limited to representing objects in closed-world settings, restricted to only operating on the datasets on which they trained. 

\begin{figure}
    \centering
    \includegraphics[width=\linewidth]{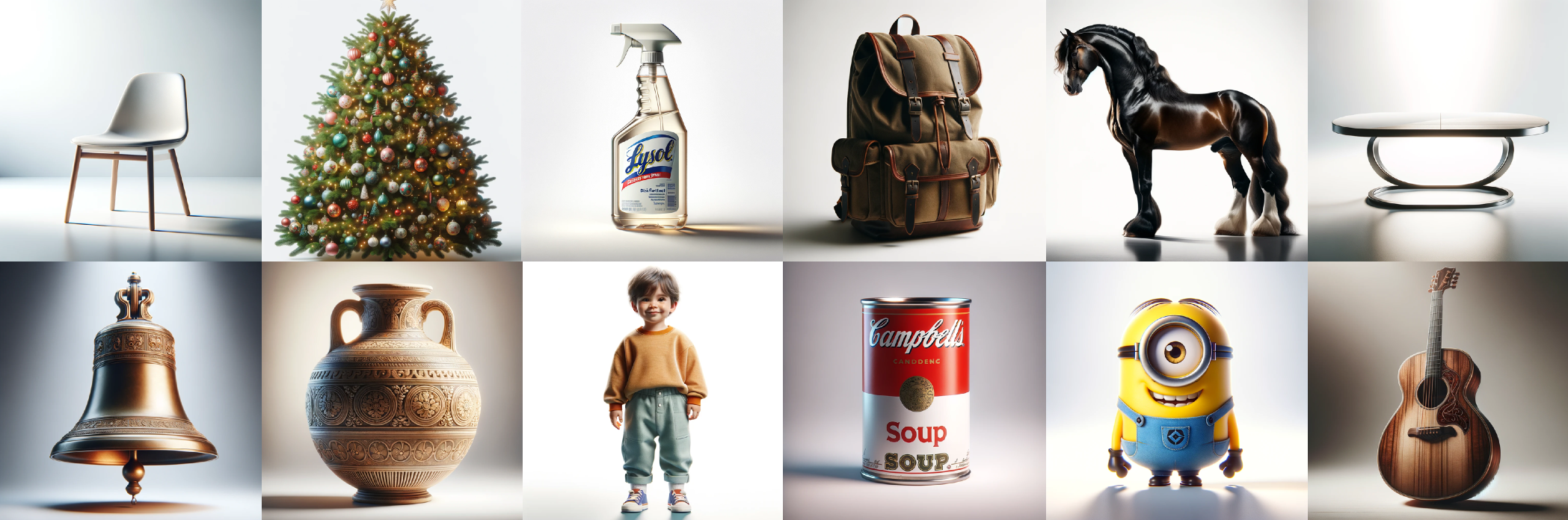}
    \caption{\textbf{Whole Objects}. Pre-trained diffusion models are able to generate all kinds of whole objects. We show samples conditioned on a category from Stable Diffusion. We leverage this synthesis ability for zero-shot  amodal reconstruction and segmentation.\vspace{-1em}}
    \label{fig:whole}
\end{figure}

In this paper, we propose an approach for zero-shot amodal segmentation and reconstruction by learning to synthesize whole objects first. Our approach capitalizes on denoising diffusion models \cite{ho2020denoising}, which are excellent representations of the natural image manifold and capture all different types of whole objects and their occlusions. Due to their large-scale training data, we hypothesize such pre-trained models have implicitly learned amodal representations (Figure \ref{fig:whole}), which we can reconfigure to encode object grouping and perform amodal completion. By learning from a synthetic dataset of occlusions and their whole counterparts, we create a conditional diffusion model that, given an RGB image and a point prompt, generates whole objects behind occlusions and other obstructions.

Our main result is showing that we are able to achieve state-of-the-art amodal segmentation results in a zero-shot setting, outperforming the methods that were specifically supervised on those benchmarks.  We furthermore show that our method can be used as a drop-in module to significantly improve the performance of existing object recognition and 3D reconstruction methods in the presence of occlusions. An additional benefit of the diffusion framework is that it allows sampling several variations of the reconstruction, naturally handling the inherent ambiguity of the occlusions.

\section{Related Work}
\label{sec:related}

We briefly review related work in amodal completion, analysis by synthesis, and denoising diffusion models for vision.

\subsection{Amodal Completion and Segmentation} \label{sec:related:completion_segmentation}
In this work, we define amodal completion as the task of generating the image of the whole object~\cite{ehsani2018segan,zhan2020self}, amodal segmentation as generating the segmentation mask of the whole object~\cite{zhu2016semantic,ke2021deep,qi2019amodal,reddy2022walt, ling2020variational}, and amodal detection as predicting the bounding box of the whole object~\cite{kar2015amodal, hsieh2023tracking}.
Most prior work focuses on the latter two tasks, due to the challenges in generating the (possibly ambiguous) pixels behind an occlusion.
In addition, to our knowledge, all prior work on these tasks is limited to a small closed-world of objects~\cite{zhan2020self,ling2020variational,ke2021deep,qi2019amodal,kar2015amodal} or to synthetic data~\cite{ehsani2018segan}.
For example, PCNet~\cite{zhan2020self}, the previous state-of-the-art method for amodal segmentation, operates only on a closed-world set of classes in Amodal COCO~\cite{zhu2016semantic}. 

Our approach, by contrast, provides rich image completions with accurate masks, generalizing to diverse zero-shot settings, while still outperforming state-of-the-art methods in a closed-world.
To achieve this degree of generalization, we capitalize on large-scale diffusion models, which implicitly learn internal representations of whole objects. We propose to unlock this capability by fine-tuning a diffusion model on a synthetically generated, realistic dataset of varied occlusions. % Our work is concurrent with \cite{xu2023amodal} and \cite{Zhan23}.

\begin{figure*}
\centering
    \includegraphics[width=0.9\textwidth]{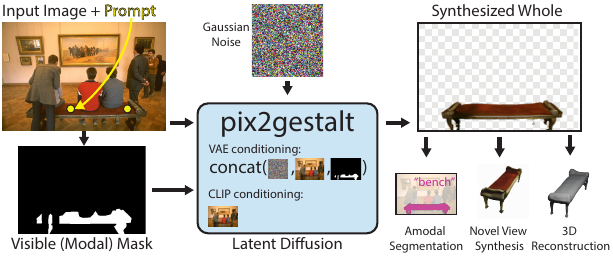}
\caption{\textbf{pix2gestalt} is an amodal completion model using a latent diffusion architecture. Conditioned on an input occlusion image and a region of interest, the whole (amodal) form is synthesized, thereby allowing other visual tasks to be performed on it too. For conditioning details, see section \ref{method:diffusion}.}
\label{fig:schematic}
\end{figure*}

% ACHAL: I think we don't need to get into the details of how we differ from Self-supervised de-occlusion, since we are drastically different, and we outperform them anyway.
% Comparison with Self-Supervised De-occlusion: 
% Supervision: All numbers from them are in fact supervised on Amodal COCO. What they mean by self-supervision is not in terms of dataset, but regarding amodal annotations:
% They train on Amodal COCO with modal annotations, learning to predict depth ordering of the scene. They use depth ordering to retrieve occluder masks, which serves as signal for amodal whole. This modal-to-amodal conversion is their self-supervision.
% Inputs:
% Depth ordering between all entities in the scene (either predicted or GT),
% Modal (visible) masks for all entities,
% Bounding boxes of all entities,
% Corresponding class labels (their Amodal COCO model specifically doesn’t use this)
% Strictly Closed World:
% Their method only works on the dataset and categories it was trained on. As such, for every dataset the report on (e.g. COCOA, KINS, LVIS), they have a separate model.
% To apply it to an image outside of Amodal COCO, we would:
% (1) estimate the depth ordering between all related objects in the scene,
% (2) use a detector to get bounding boxes,
% (3) for their models besides amodal coco-trained, we would also need object labels (mapping to the training dataset of specific pre-trained version).

\subsection{Analysis by Synthesis}\label{related:recognition}
Our approach is heavily inspired by analysis by synthesis~\cite{yuille2006vision} -- a generative approach for visual reasoning. Image parsing~\cite{tu2005image} was a representative work that unifies segmentation, recognition, and detection by generation. Prior works have applied the analysis by synthesis approaches on various problems including face recognition~\cite{tu2005image,blanz2023morphable}, pose estimation~\cite{zhang20233d,ma2022robust}, 3D reconstruction~\cite{liu2023humans,liu2022shadows}, semantic image editing~\cite{abdal2019image2stylegan,liu2023landscape,zhu2020domain}. In this paper, we aim to harness the power of generative models trained with internet-scale data for the task of amodal completion, thereby aiding various tasks such as recognition, segmentation, and 3D reconstruction in the presence of occlusions.

\subsection{Diffusion Models} \label{related:diffusion}
Recently, Denoising Diffusion Probabilistic Model~\cite{ho2020denoising}, or DDPM, has emerged as one of the most widely used generative architectures in computer vision due to its ability to model multi-modal distributions, training stability, and scalability. \cite{dhariwal2021diffusion} first showed that diffusion models outperform GANs~\cite{goodfellow2014generative} in image synthesis. Stable Diffusion~\cite{rombach2022high}, trained on LAION-5B~\cite{schuhmann2022laion}, applied diffusion model in the latent space of a variational autoencoder~\cite{kingma2013auto} to improve computational efficiency. Later, a series of major improvements were made to improve diffusion model performance~\cite{ho2022classifier,song2020denoising}. With the release of Stable Diffusion as a strong generative prior, many works have adapted it to solve tasks in different domain such as image editing~\cite{brooks2022instructpix2pix,gal2022image,ruiz2023dreambooth}, 3D~\cite{liu2023zero, deitke2023objaverse, wu2023sin3dm}, and modal segmentation~\cite{xu2023odise,amit2021segdiff,baranchuk2021label}. In this work, we leverage the strong occlusion and complete object priors provided by internet-pretrained diffusion model to solve the zero-shot amodal completion task.
\section{Amodal Completion via Generation} \label{method}

Given an RGB image $x$ with an occluded object that is partially visible, our goal is to predict a new image with the shape and appearance of the whole object, and only the whole object. Our approach will accept any point or mask as a prompt $p$ indicating the modal object:
\begin{align*}
    \hat{x}_p = f_\theta(x, p)
\end{align*}
where $\hat{x}_p$ is our estimate of the whole object indicated by $p$. Mapping from $x$ to this unified whole form, \ie\textit{gestalt} of the occluded object, we name our method \textbf{pix2gestalt}. We want $\hat{x}$ to be perceptually similar to the true but unobserved whole of the object as if there was no occlusion. We will use a conditional diffusion model (see Figure~\ref{fig:schematic}) for $f_\theta$.

An advantage of this approach is that, once we estimate an image of the whole object $\hat{x}$, we are able to perform any other computer vision task on it, providing a unified method to handle occlusions across different tasks. Since we will directly synthesize the pixels of the whole object, we can aid off-the-shelf approaches to perform segmentation, recognition, and 3D reconstruction of occluded objects. 

To perform amodal completion, $f$ needs to learn a representation of whole objects in the visual world. Due to their scale of training data, we will capitalize on large pretrained diffusion models, such as Stable Diffusion, which are excellent representations of the natural image manifold and have the support to generate unoccluded objects. However, although they generate high-quality images, their representations do not explicitly encode the grouping of objects and their boundaries to the background. 

\subsection{Whole-Part Pairs}\label{method:dataset}

To learn the conditional diffusion model $f$ with the ability for grouping, we build a large-scale paired dataset of occluded objects and their whole counterparts. Unfortunately, collecting a natural image dataset of these pairs is challenging at scale. Prior datasets provide amodal segmentation annotations \cite{zhu2016semantic,qi2019amodal}, but they do not reveal the pixels behind an occlusion.  Other datasets have relied on graphical simulation \cite{HuCVPR2019}, which lack the realistic complexity and scale of everyday object categories.    

\begin{figure*}[t]
\centering\includegraphics[width=0.9\textwidth]{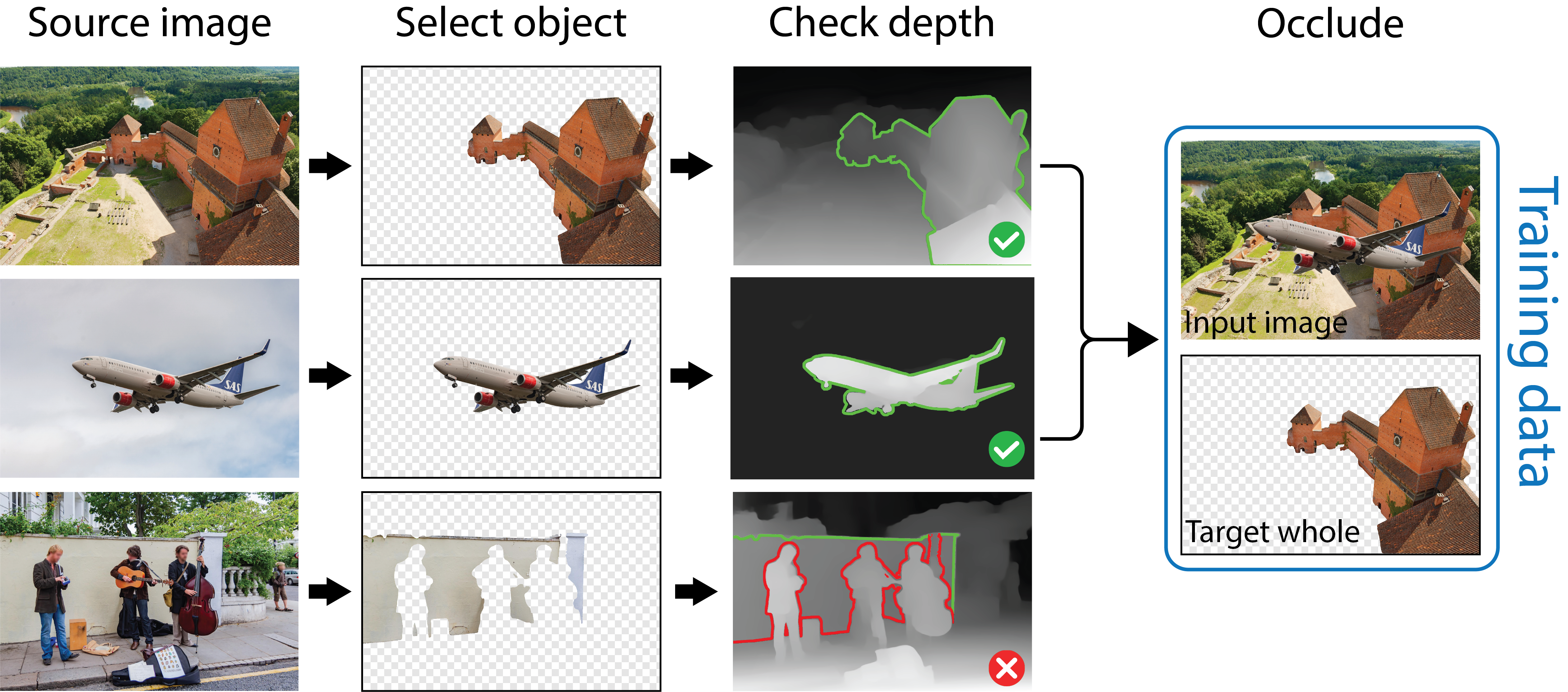}
\caption{\textbf{Constructing Training Data.} To ensure we only occlude whole objects, we use a heuristic that objects closer to the camera than its neighbors are likely whole objects. The \textcolor{green}{green} outline around the object shows where the estimated depth is closer to the camera than the background (the \textcolor{red}{red} shows when it is not).}
\label{fig:dataset}
\end{figure*}

We build paired data by automatically overlaying objects over natural images. The original images provide ground-truth for the content behind occlusions. However, we need to ensure that we only occlude whole objects in this construction, as otherwise our model could learn to generate incomplete objects. To this end, we use a heuristic that, if the object is closer to the camera than its neighboring objects, then it is likely a whole object.  We use Segment Anything~\cite{kirillov2023segment} to automatically find object candidates in the SA-1B dataset, and use the off-the-shelf monocular depth estimator MiDaS~\cite{birkl2023midas} to select which objects are whole.
For each image with at least one whole object, we sample an occluder and superimpose it, resulting in a paired dataset of $837K$ images and their whole counterparts. Figure \ref{fig:dataset} illustrates this construction and shows examples of the heuristic.

\subsection{Conditional Diffusion}\label{method:diffusion}
Given pairs of an image $x$ and its whole counterpart $\hat{x}_p$, we fine-tune a conditional diffusion model to perform amodal completion while maintaining the zero-shot capabilities of the pre-trained model.
We solve for the following latent diffusion objective:
\begin{align*}
   \min_{\theta}\;  \mathbb{E}_{z \sim \mathcal{E}(x), t, \epsilon \sim \mathcal{N}(0, 1)} \left[ ||\epsilon - \epsilon_{\theta}(z_t, \mathcal{E}(x), t, \mathcal{E}(p), \mathcal{C}(x))||_2^2 \right]
\end{align*}
where $0 \le t < 1000$ is the diffusion time step, $z_t$ is the embedding of the noised amodal target image $\hat{x}_p$. $\mathcal{C}(x)$ is the CLIP embedding of the input image, and $\mathcal{E}(\cdot)$ is a VAE embedding. Following \cite{brooks2022instructpix2pix, liu2023zero}, we apply classifier-free guidance (CFG) ~\cite{ho2022classifier} by setting the conditional information to a null vector randomly.

Amodal completion requires reasoning about the whole shape, its appearance, and contextual visual cues of the scene. We adapt the design in ~\cite{brooks2022instructpix2pix,liu2023zero} to condition the diffusion model~$\epsilon_\theta$ in two separate streams. $\mathcal{C}(x)$ conditions the diffusion model $\epsilon_\theta$ via cross-attention on the semantic features of the partially visible object in $x$ as specified by $p$, providing high-level perception. On the VAE stream, we channel concatenate $\mathcal{E}(x)$ and $z_t$, providing low-level visual details (shade, color, texture), as well as $\mathcal{E}(p)$ to indicate the visible region of the object. 

After $\epsilon_\theta$ is trained, $f$ can generate $\hat{x}_p$ by performing iterative denoising  \cite{rombach2022high}. The CFG can be scaled to control impact of the conditioning on the completion.

\subsection{Amodal Base Representations} \label{related:base}

Since we synthesize RGB images of the whole object, our approach makes it straightforward to equip various computer vision methods with the ability to handle occlusions. We discuss a few common cases.

\textbf{Image Segmentation} aims to find the spatial boundaries of an object given an image $x$ and an initial prompt $p$. We can perform amodal segmentation by completing an occluded object with $f$, then thresholding the result to obtain an amodal segmentation map. Note that this problem is under-constrained as there are multiple possible solutions. Given the uncertainty, we found that sampling multiple completions and performing a majority vote on the segmentation masks works best in practice.

\textbf{Object Recognition} is the task of classifying an object located in an bounding box or mask $p$. We can zero-shot recognize significantly occluded objects by first completing the whole object with $f$, then classifying the amodal completion with CLIP. 

\textbf{3D Reconstruction} estimates the appearance and geometry of an object. We can zero-shot reconstruct objects with partial occlusions by first completing the whole object with $f$, then applying SyncDreamer and Score Distillation Sampling~\cite{poole2022dreamfusion} to estimate a textured mesh.

\section{Experiments} 
We evaluate pix2gestalt's ability to perform zero-shot amodal completion for three tasks: amodal segmentation, occluded object recognition, and amodal 3D reconstruction.
We show that our method provides amodal completions that directly lead to strong results in all tasks. 

\begin{figure*}[t]
\centering \includegraphics[width=0.95\textwidth]{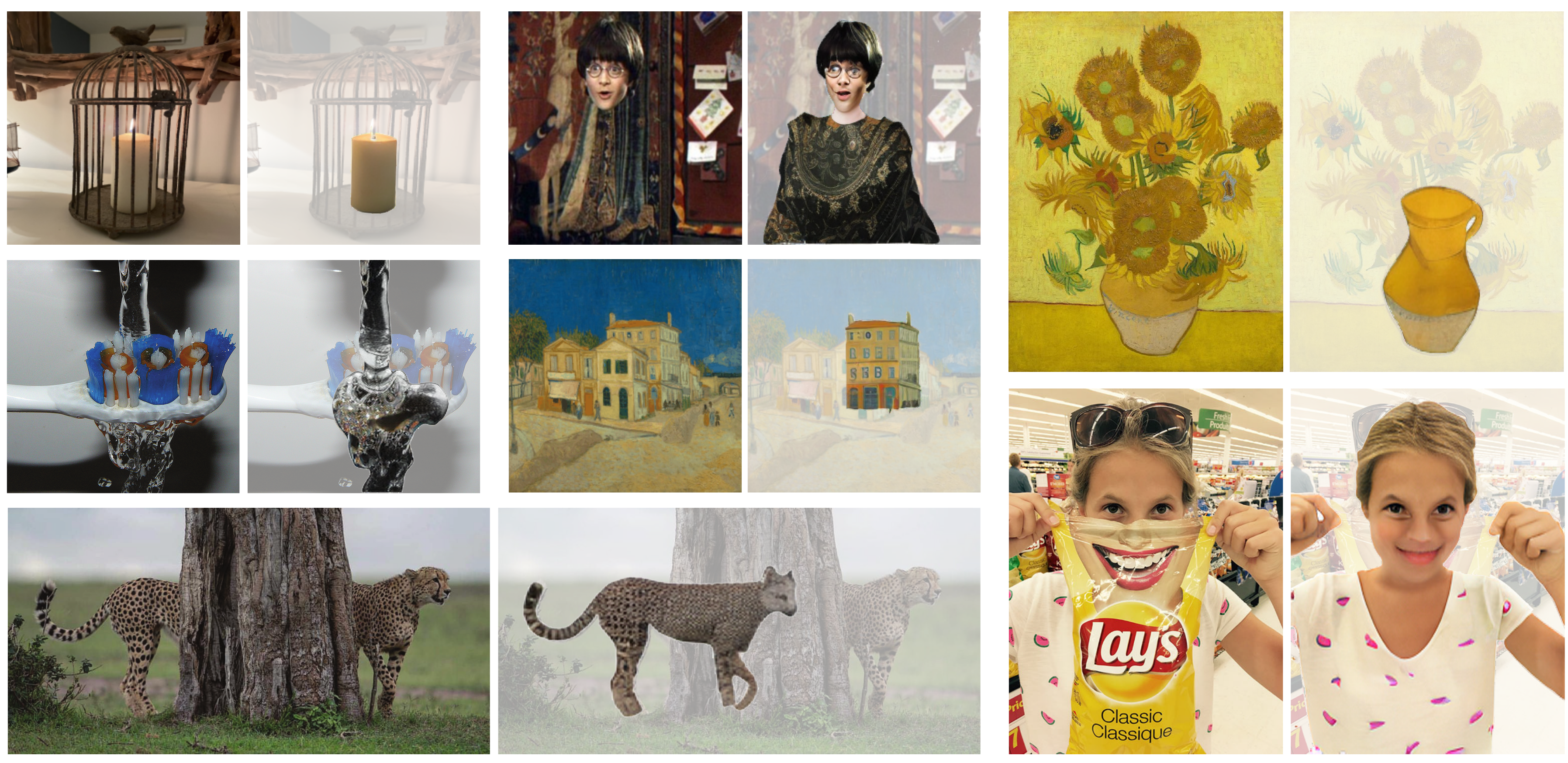}
\caption{\textbf{In-the-wild Amodal Completion and Segmentation.} We find that pix2gestalt is able to synthesize whole objects in novel situations, including artistic pieces, images taken by an iPhone, and illusions.}
\label{fig:ood}
\end{figure*}

\begin{figure*}
\centering
    \includegraphics[width=0.95\textwidth]{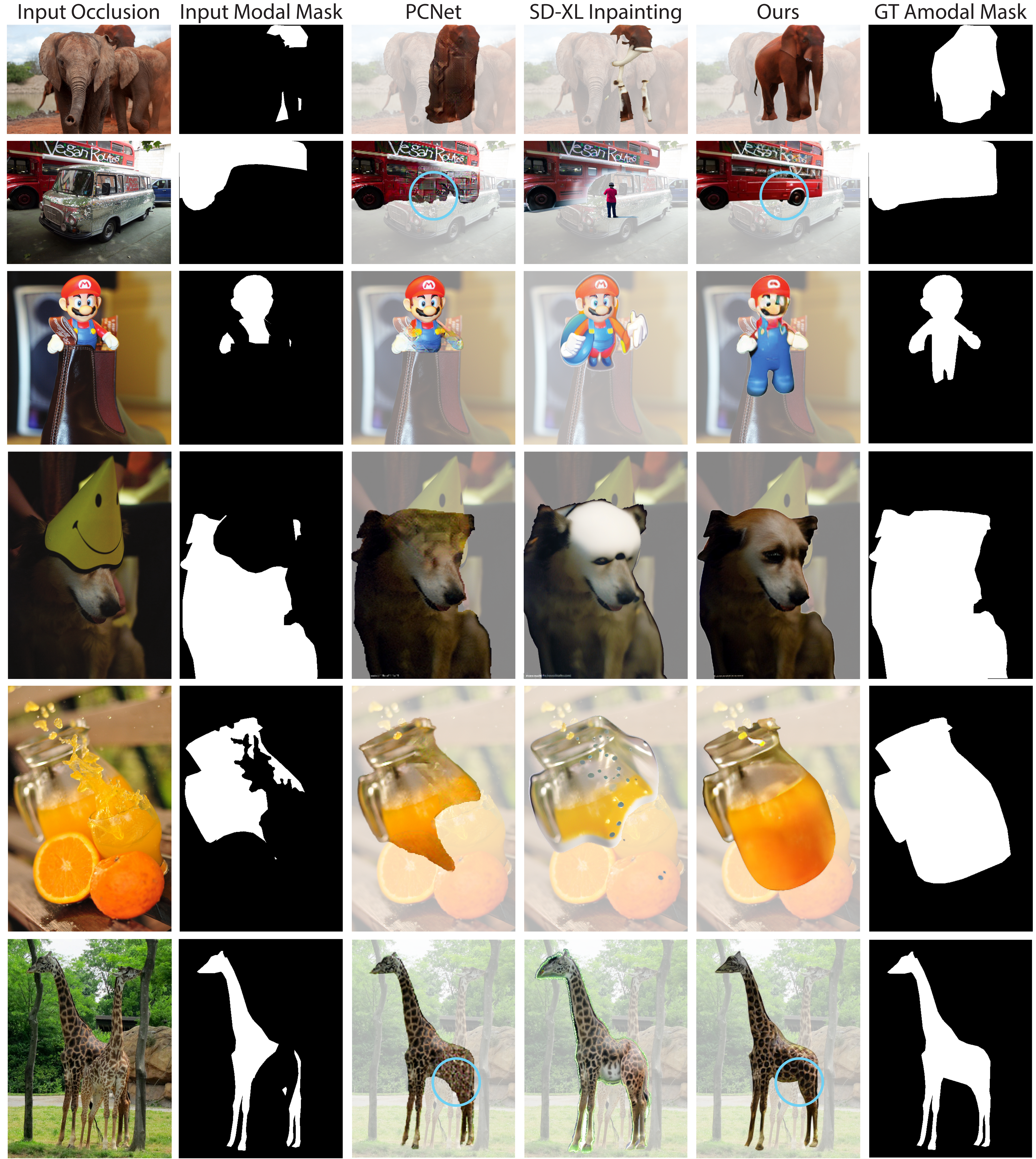}
\caption{\textbf{Amodal Completion and Segmentation Qualitative Results on Amodal COCO}. In \textcolor{cyan}{blue} circles, we highlight completion regions that, upon a closer look, have a distorted texture in the PCNet baseline, and a correct one in our results.}
\label{fig:amodal_qualitative}
\end{figure*}

\subsection{Amodal Segmentation}
\label{sec:segmentation}
\textbf{Setup.}
Amodal segmentation requires segmenting the full extent of a (possibly occluded) object.
We evaluate this task on the Amodal COCO (COCO-A)~\cite{zhu2016semantic} and Amodal Berkeley Segmentation (BSDS-A) datasets~\cite{MartinFTM01}.
For evaluation, COCO-A provides 13,000 amodal annotations of objects in 2,500 images, while BSDS-A provides 650 objects from 200 images.
For both datasets, we evaluate methods that take as input an image and a (modal) mask of the visible extent of an object, and output an amodal mask of the full-extent of the object.
Following~\cite{zhan2020self}, we evaluate segmentations using mean intersection-over-union (mIoU).
We follow the strategy in \Cref{related:base} to convert our amodal completions into segmentation masks.

We evaluate three baselines for amodal segmentation.
The first method is PCNet~\cite{zhan2020self}, which is trained for amodal segmentation specifically for COCO-A.
Next, we compare to two zero-shot methods, which do not train on COCO-A: 
Segment Anything (SAM)~\cite{kirillov2023segment}, a strong \textit{modal} segmentation method, and Inpainting using Stable Diffusion-XL \cite{podell2023sdxl}.
To evaluate inpainting methods, we provide as input an image with all but the visible object region erased, and convert the completed image output by the method into an amodal segmentation mask following the same strategy as for our method.

\textbf{Results.}
Table \ref{tab:amodalseg} compares pix2gestalt with prior work.
Despite never training on the COCO-A dataset, our method outperforms all baselines, including PCNet, which uses COCO-A images for training, and even PCNet-Sup, which is supervised using human-annotated amodal segmentations from COCO-A's training set. Compared to other zero-shot methods, our improvements are dramatic, validating the generalization abilities of our method. Notably, we also outperform the inpainting baseline which is based off a larger, more recent variant of Stable Diffusion \cite{podell2023sdxl}. This demonstrates that internet-scale training alone is not sufficient and our fine-tuning approach is key to reconfigure priors from pre-training for amodal completion.

\begin{figure*}
\centering \includegraphics[width=\textwidth]{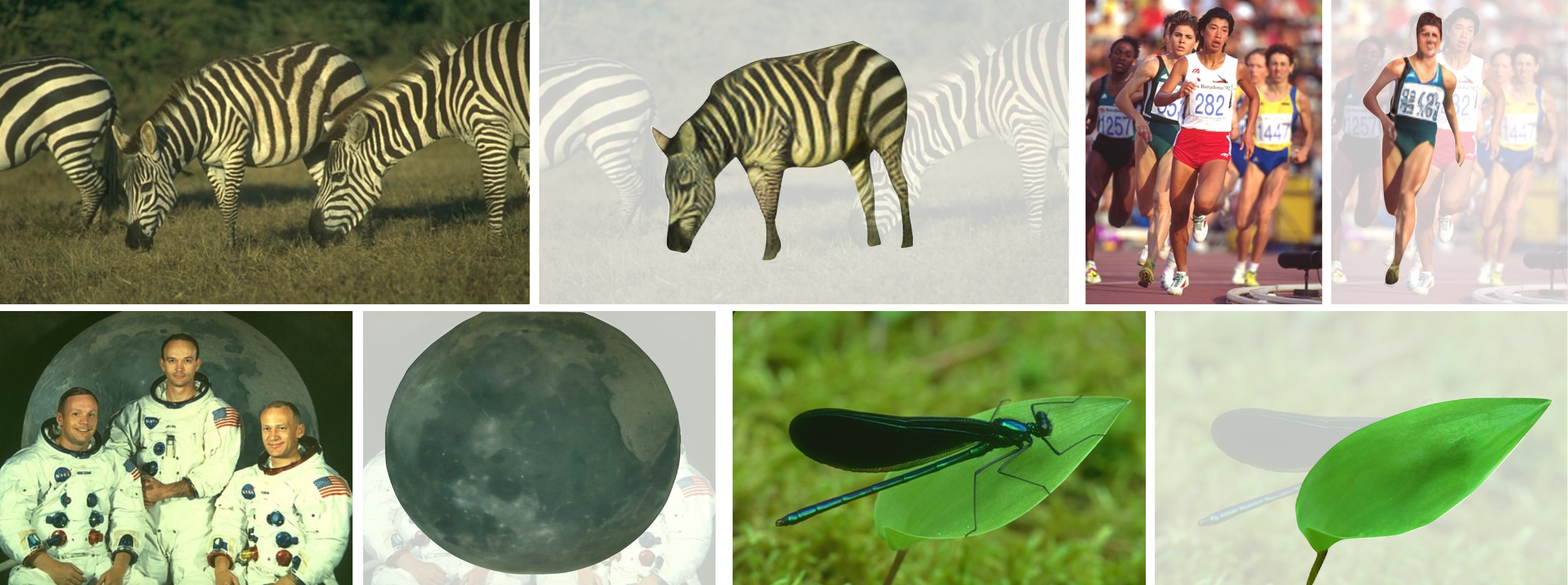} \caption{\textbf{Amodal Berkeley Segmentation Dataset Qualitative Results.} Our method provides accurate, complete reconstructions of occluded objects.}
\label{fig:amodalbsds}
\end{figure*}

We further analyze amodal completions qualitatively in Figure ~\ref{fig:amodal_qualitative}. While SD-XL often hallucinates extraneous, unrealistic details (e.g. person in front of the bus in the second row), PCNet tends to fail to recover the full extent of objects---often only generating the visible region, as in the Mario example in the third row. In contrast, pix2gestalt provides accurate, complete reconstructions of occluded objects on both COCO-A (Figure ~\ref{fig:amodal_qualitative}) and BSDS-A (Figure ~\ref{fig:amodalbsds}). 
Our method generalizes well beyond the typical occlusion scenarios found in those benchmarks. Figure~\ref{fig:ood} shows several examples of out-of-distribution images, including art pieces, illusions, and images taken by ourselves that are successfully handled by our method. Note that no prior work has shown open-world generalization (see \ref{sec:related:completion_segmentation}).

\begin{figure}[t]
\centering
    \includegraphics[width=\columnwidth]{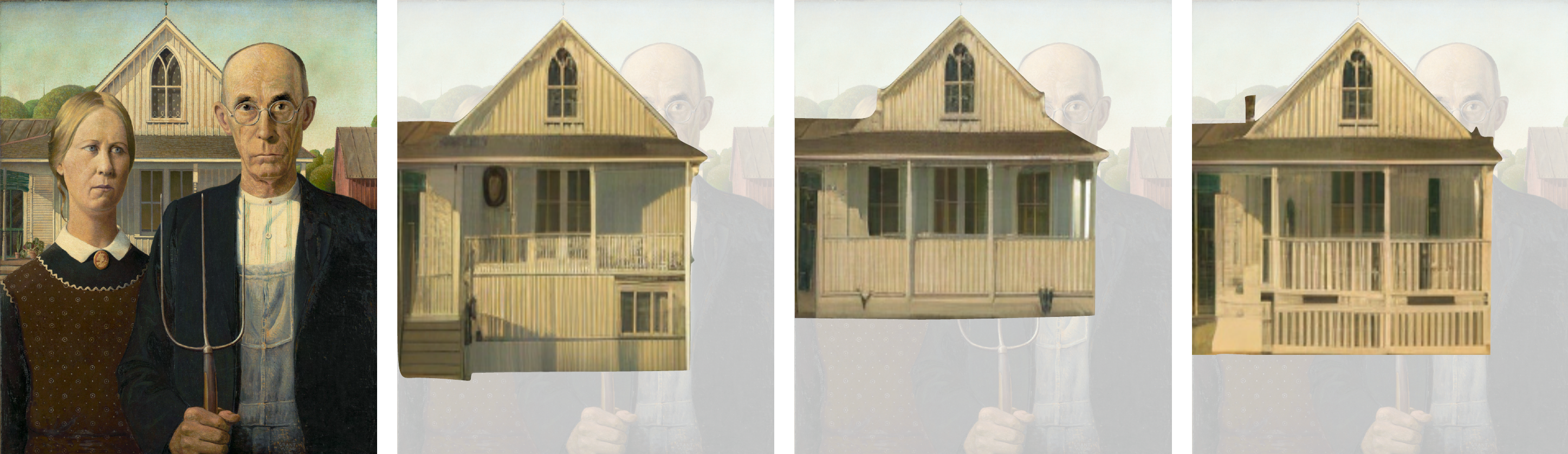}
\caption{\textbf{Diversity in Samples}. Amodal completion has inherent uncertainties. By sampling from the diffusion process multiple times, the method synthesizes multiple plausible wholes that are consistent with the input observations.}
\label{fig:uncertainty}
\end{figure}

\begin{figure}
    \centering
    \includegraphics[width=\columnwidth]{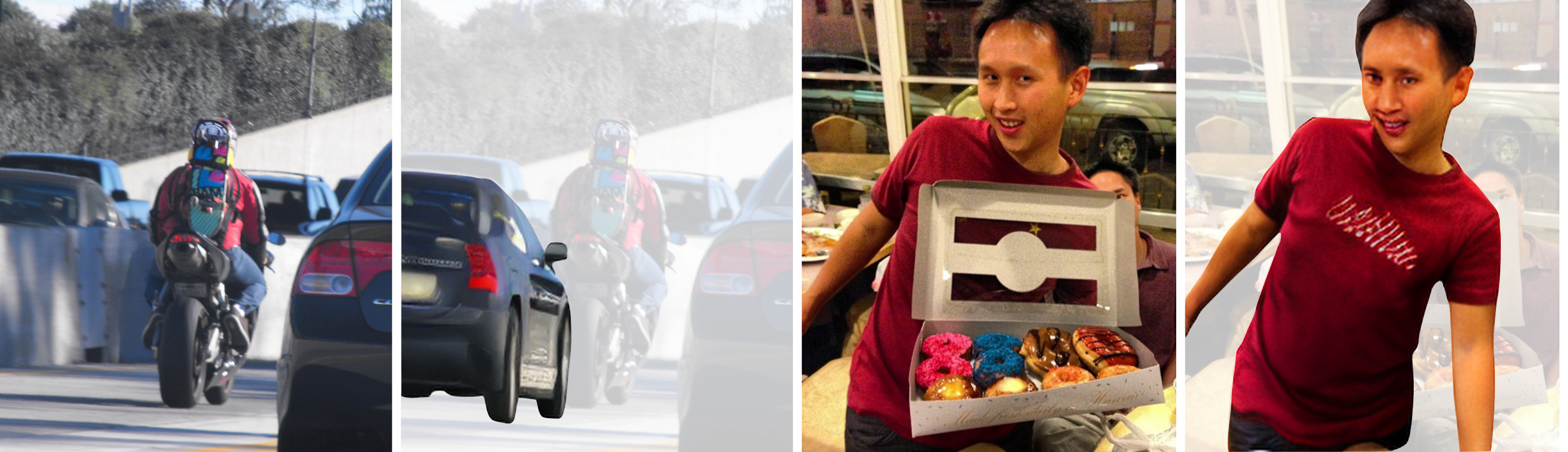}
    \caption{\textbf{Common-sense and Physics Failures.} Left: reconstruction has the car going in the wrong direction. Right: reconstruction contradicts physics, failing to capture that a hand must be holding the donut box.}
    \label{fig:failures}
\end{figure}

Figure~\ref{fig:uncertainty} illustrates the ability of the approach to generate diverse samples in shape and appearance when there is uncertainty in the final completion. For example, it is able to synthesize several plausible completions of the occluded house in the painting. We quantitatively evaluate the diversity of our samples in the last row of Table~\ref{tab:amodalseg} by sampling from our model three times and reporting the performance for the best sample (``Best of 3''). 
%Our approach creates sufficiently diverse and high-quality samples that we obtain significant gains of over four points from just three samples.
Finally, we found limitations of our approach in situations that require commonsense or physical reasoning. We show two examples in Figure \ref{fig:failures}.

\begin{table}[t]
    \setlength{\tabcolsep}{2pt}
    \caption{\textbf{Amodal Segmentation Results}. We report \textbf{mIoU (\%)} $\uparrow$ on Amodal COCO \cite{zhu2016semantic} and on Amodal Berkeley Segmentation Dataset \cite{zhu2016semantic, MartinFTM01}. $^*$PCNet-Sup trains using ground truth amodal masks from COCO-Amodal. See \Cref{sec:segmentation} for analysis.} 
    \label{tab:amodalcoco}
    \centering
    \begin{tabular}{clcc}
        \toprule
        {Zero-shot} & Method & COCO-A & BSDS-A \\ 
        \midrule
        \ding{55} & PCNet\phantom{$^{*}$}~\cite{zhan2020self} & 81.35 & - \\
        \ding{55} & PCNet-Sup$^{*}$~\cite{zhan2020self} & 82.53\makebox[0pt]{\;\;$^*$} & - \\
        \midrule
        \checkmark & SAM~\cite{kirillov2023segment} & 67.21 & 65.25 \\
        % \checkmark & Fine-Tuned SAM & 75.32 & 76.21 \\
        \checkmark & SD-XL Inpainting~\cite{podell2023sdxl} & 76.52 & 74.19 \\
        
        % \ding{55} & PCNet \cite{zhan2020self} & 76.91 & - \\
        \checkmark & {Ours} & \textbf{82.87} & \textbf{80.76} \\
        \midrule
        \checkmark & {Ours: Best of 3} & \textbf{87.10} & \textbf{85.68} \\
        \bottomrule
    \end{tabular}
    \label{tab:amodalseg}
\end{table}

\subsection{Occluded Object Recognition} \label{sec:recognition} Next, we evaluate the utility of our method for recognizing occluded objects.

\textbf{Setup.} We use the Occluded and Separated COCO benchmarks~\cite{zhan2022triocc} for evaluating classification accuracy under occlusions. The former consists of partially occluded objects, whereas Separated COCO contains objects whose modal region is separated into disjoint segments by the occluder(s), resulting in a more challenging problem setting. We evaluate on all 80 COCO semantic categories in the datasets using Top 1 and Top 3 accuracy.

We use CLIP~\cite{radford2021learning} as the base open-vocabulary classifier. As baselines, we evaluate CLIP without any completion, reporting three variants: providing the entire image (CLIP), providing the entire image with a visual prompt (a red circle, as in Shtedritski \etal~\cite{shtedritski2023does}) around the occluded object, or providing an image with all but the visible portion of the occluded object masked out.
To evaluate our approach, we first utilize it to complete the occluded object, and then classify the output image using CLIP.

\textbf{Results.} Table \ref{tab:recognition} compares our method with the baselines. Visual prompting with a red circle (RC) and masking all but the visible object (Vis.~Obj.) provide improvements over directly passing the image to CLIP on the simpler Occluded COCO benchmark, but fail to improve, and some times even decreases the performance of the baseline CLIP on the more challenging Separated COCO variant.
Our method (Ours + CLIP), however, strongly outperforms all baselines for both the occluded and separated datasets, verifying the quality of our completions.

\begin{table}[t]
\setlength{\tabcolsep}{2pt} \caption{\textbf{Occluded Object Recognition}. We report zero-shot classification accuracy on Occluded and Separated COCO \cite{zhan2022triocc}. While simple baselines fail to improve CLIP performance in the more challenging Separated COCO setting, our method consistently improves recognition accuracy by large margins. See \Cref{sec:recognition} for analysis.}
    \label{tab:recognition}
    \centering
    \begin{tabular}{l c c c c c} 
        \toprule
        Method & \multicolumn{2}{c}{{Top 1 Acc. (\%) $\uparrow$}} & \multicolumn{2}{c}{{Top 3 Acc. (\%) $\uparrow$}} \\
        \cmidrule(lr){2-3} \cmidrule(lr){4-5} 
        & Occluded & Sep. & Occluded & Sep.\\
        \midrule
        
        CLIP \cite{radford2021learning} & 23.33 & 26.04 & 43.84 & 43.19 \\
        CLIP + RC \cite{shtedritski2023does} & 23.46 & 25.64 & 43.86 & 43.24 \\
        Vis. Obj. + CLIP & 34.00 & 21.10 & 49.26 & 34.70 \\
        {Ours + CLIP} & \textbf{43.39} & \textbf{31.15} & \textbf{58.97} & \textbf{45.77} \\
        \bottomrule
    \end{tabular}
\end{table}

\begin{figure*}
\centering
\includegraphics[width=0.95\textwidth]{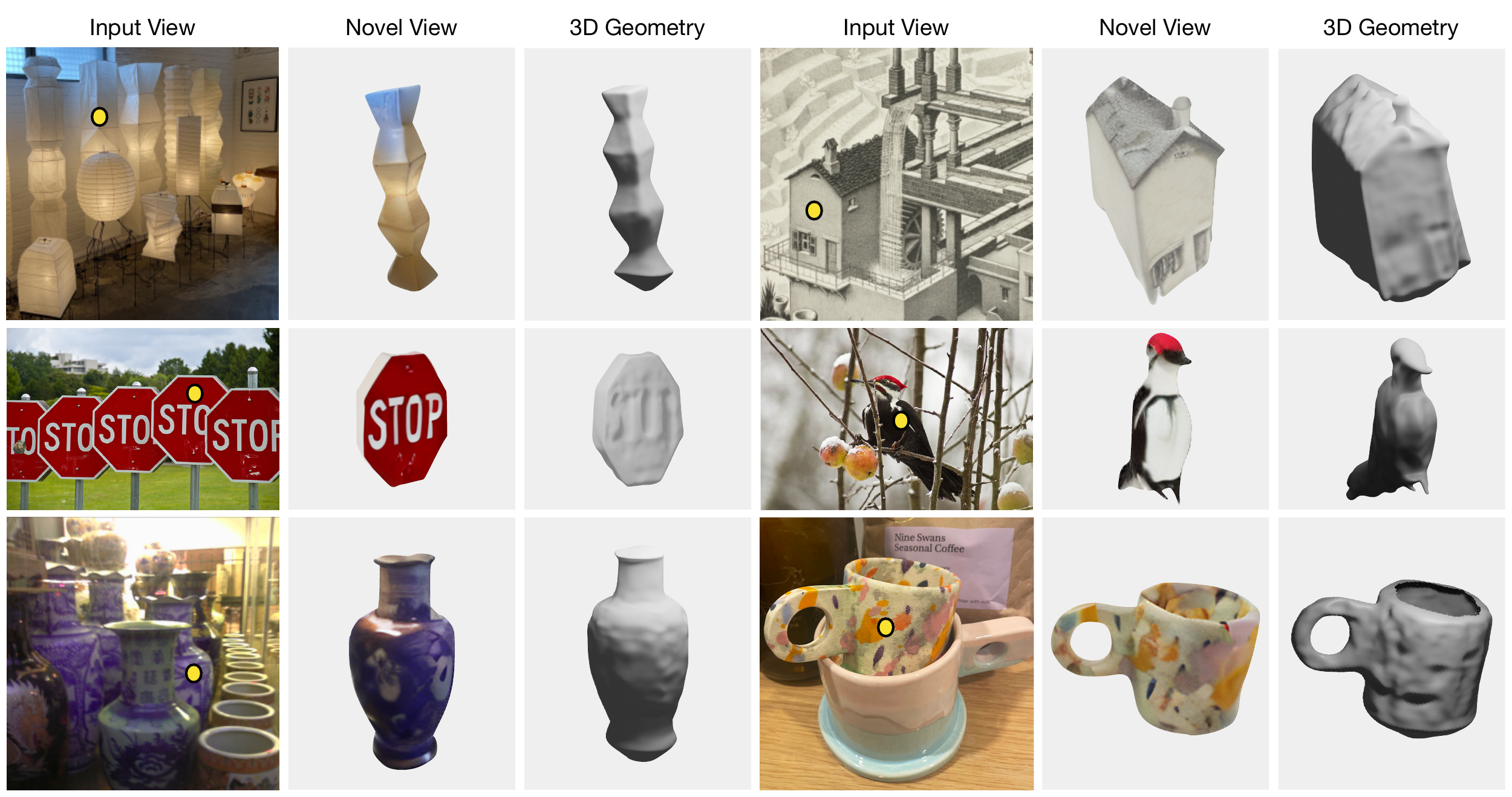}
\caption{\textbf{Amodal 3D Reconstruction qualitative results}. The object of interest is specified by a point prompt, shown in yellow. Incorporating pix2gestalt as a drop-in module to state-of-the-art 3D reconstruction models allows us to address challenging and diverse occlusion scenarios with ease.}
\label{fig:three_d} 
\end{figure*}
\subsection{Amodal 3D Reconstruction}
\label{sec:3d_reconstruction}

Finally, we evaluate our method for improving 3D reconstruction of occluded objects.

\textbf{Setup.} We focus on two tasks: Novel-view synthesis and single-view 3D reconstruction.

To demonstrate pix2gestalt's performance as a drop-in module to 3D foundation models \cite{liu2023zero, liu2023syncdreamer, zeronvs}, we replicate the evaluation procedure of Zero-1-to-3~\cite{liu2023syncdreamer, liu2023zero} on Google Scanned Objects (GSO) \cite{downs2022google}, a dataset of common household objects 3D scanned for use in embodied, synthetic, and 3D perception tasks.
We use 30 randomly sampled objects from GSO ranging from daily objects to animals.
For each object, we render a 256x256 image with synthetic occlusions sampled from the full dataset of 1,030 objects in GSO.
We render from a randomly sampled view to avoid canonical poses, and generate two occluded images for each of the 30 objects, resulting in 60 samples.

For amodal novel-view synthesis, we quantitatively evaluate our method using 3 metrics: PSNR, SSIM \cite{wang2004image}, and LPIPS \cite{zhang2018perceptual}, measuring the image-similarity of the input and ground truth views. For 3D reconstruction, we use the Volumetric IoU and Chamfer Distance metrics.
We compare our approach with SyncDreamer \cite{liu2023syncdreamer}, a 3D generative model that fine-tunes Zero123-XL \cite{deitke2023objaverse, liu2023zero} for multi-view consistent novel view synthesis and consequent 3D reconstruction with NeuS \cite{wang2021neus} and NeRF \cite{mildenhall2020nerf}. 
Our first baseline provides as input to SyncDreamer the segmentation mask of all foreground objects, following the standard protocol.
To avoid reconstructing occluded objects, we additionally evaluate two variants that use SAM~\cite{kirillov2023segment} to estimate the mask of only the object of interest, or the ground truth mask for the object of interest (GT Mask).
Finally, to evaluate our method, we provide as input the full object completed by our method, along with the corresponding amodal mask.
We evaluate two variants of our method: One where we provide a modal mask for the object of interested as estimated by SAM (Ours (SAM Mask)) and one where we use the ground truth modal mask (Ours (GT Mask)).
% We evaluate SyncDreamer on the rendered images  where the alpha channel specifies the entirety of the foreground object (union of the occluder and modal mask of the object of interest) - without any further cues about the occluder.

%Then for a more fair comparison, as our second baseline, we add SAM \cite{kirillov2023segment} as modal pre-processor module to specify the region of interest, i.e. which pixels correspond to the visible region of the object.
%For this baseline, the alpha channel represents the modal region only, hence the 3D model is prompted to reconstruct only the visible region. It does not know any better about the full extent of the object without an amodal completion and segmentation module. We repeat this as our fourth baseline with ground truth modal masks, to show an upper bound modal pre-processor. Finally, we show our model as a drop-in \textit{amodal} pre-processor module by performing amodal completion and segmentation on the input image. Our predicted whole object and predicted amodal mask as inputted to SyncDreamer as the RGB and alpha channels, respectively. For a more realistic test-time comparison, we repeat this with the same SAM modal masks that were inputted to SyncDreamer.

\begin{table}[t]
    \caption{\textbf{Single-view 3D Reconstruction.} We report Chamfer Distance and Volumetric IoU for Google Scanned Objects. See \Cref{sec:3d_reconstruction} for analysis.}
    \label{tab:3drec}
    \centering
    \begin{tabular}{l c c}
        \toprule
        & \multicolumn{1}{c}{{CD} $\downarrow$} & \multicolumn{1}{c}{{IoU} $\uparrow$} \\   
        \midrule
        SyncDreamer \cite{liu2023syncdreamer} & 0.0884 & 0.2741 \\
        SAM Mask + SyncDr. & 0.1182 &  0.0926 \\
        {Ours (SAM Mask) + SyncDr.}  &  \textbf{0.0784} &  \textbf{0.3312} \\
        \midrule
        GT Mask + SyncDr.& 0.1084 &  0.1027 \\
        {Ours (GT Mask) + SyncDr.} & \textbf{0.0681} & \textbf{0.3639} \\
        \bottomrule
    \end{tabular}
\end{table}

\textbf{Results.} 
We compare our approach with the two baselines in Table~\ref{tab:nvs} for novel view synthesis and Table~\ref{tab:3drec} for 3D reconstruction. Quantitative results demonstrate that we strongly outperform the baselines for both tasks. 
In novel-view synthesis, we outperform SAM + SyncDreamer on the image reconstruction metrics, LPIPS \cite{zhang2018perceptual} and PSNR \cite{wang2004image}.  Compared to SAM as a modal pre-processor, we obtain these improvements as a drop-in module to SyncDreamer while still retaining equivalent image quality (Table~\ref{tab:nvs}, SSIM \cite{wang2004image}). With ground truth mask inputs, 
we obtain further image reconstruction gains. Moreover, even though our approach utilizes an additional diffusion step compared to SyncDreamer only, we demonstrate less image quality degradation.

For reconstruction of the 3D geometry, 
our fully automatic method outperforms all of the baselines for both volumetric IoU and Chamfer distance metrics, even the baselines that use ground masks. Providing the ground truth to our approach further improves the results. Figure \ref{fig:three_d} shows qualitative evaluation for 3D reconstruction of occluded objects, ranging from an Escher lithograph to in-the-wild images. 

\begin{table}[t]
\setlength{\tabcolsep}{3pt}
    \caption{\textbf{Novel-view synthesis from one image.} We report results on Google Scanned Objects \cite{downs2022google}. Note SSIM measures image quality, not novel-view accuracy. See \Cref{sec:3d_reconstruction} for analysis.}
    \label{tab:nvs}
    \centering
    \begin{tabular}{l c c | c}
        \toprule
        & \multicolumn{1}{c}{{LPIPS} $\downarrow$} & \multicolumn{1}{c}{{PSNR} $\uparrow$} & \multicolumn{1}{|c}{{SSIM} $\uparrow$} \\   
        \midrule
        SyncDreamer \cite{liu2023syncdreamer} & 0.3221 & 11.914 & 0.6808 \\
        SAM + SyncDr. & 0.3060 & 12.432 & {0.7248} \\

        {Ours (SAM Mask) + SyncDr.}  & \textbf{0.2848} & \textbf{13.868} & {0.7211}  \\
        \midrule
        GT Mask + SyncDr. & 0.2905 & 12.561 & {0.7322} \\
        {Ours (GT Mask) + SyncDr.} & \textbf{0.2631} & \textbf{14.657} & {0.7328} \\
        \bottomrule
    \end{tabular}
\end{table}

\section{Conclusion}
In this work, we proposed a novel approach for zero-shot amodal segmentation via synthesis. Our model capitalizes on whole object priors learned by internet-scale diffusion models and unlocks them via fine-tuning on a synthetically generated dataset of realistic occlusions. We then demonstrated that synthesizing the whole object makes it straightforward to equip various computer vision methods with the ability to handle occlusions. In particular, we reported state-of-the art results on several benchmarks for amodal segmentation, occluded object recognition and 3D reconstruction. 

\textbf{Acknowledgements:} This research is based on work partially supported by the Toyota Research Institute, the DARPA MCS program under Federal Agreement No. N660011924032, the NSF NRI Award \#1925157, and the NSF AI Institute for Artificial and Natural Intelligence Award \#2229929. DS is supported by the Microsoft PhD Fellowship.
% \input{sec/5_futurework}
% \input{sec/6_conclusion}

% \newpage
{
    \small
    \bibliographystyle{ieeenat_fullname}
    \bibliography{main}

\begin{thebibliography}{53}
\providecommand{\natexlab}[1]{#1}
\providecommand{\url}[1]{\texttt{#1}}
\expandafter\ifx\csname urlstyle\endcsname\relax
  \providecommand{\doi}[1]{doi: #1}\else
  \providecommand{\doi}{doi: \begingroup \urlstyle{rm}\Url}\fi

\bibitem[Abdal et~al.(2019)Abdal, Qin, and Wonka]{abdal2019image2stylegan}
Rameen Abdal, Yipeng Qin, and Peter Wonka.
\newblock {Image2StyleGAN}: How to embed images into the stylegan latent space?
\newblock In \emph{ICCV}, 2019.

\bibitem[Amit et~al.(2021)Amit, Shaharbany, Nachmani, and Wolf]{amit2021segdiff}
Tomer Amit, Tal Shaharbany, Eliya Nachmani, and Lior Wolf.
\newblock Segdiff: Image segmentation with diffusion probabilistic models.
\newblock \emph{arXiv preprint arXiv:2112.00390}, 2021.

\bibitem[Baranchuk et~al.(2021)Baranchuk, Rubachev, Voynov, Khrulkov, and Babenko]{baranchuk2021label}
Dmitry Baranchuk, Ivan Rubachev, Andrey Voynov, Valentin Khrulkov, and Artem Babenko.
\newblock Label-efficient semantic segmentation with diffusion models.
\newblock \emph{arXiv preprint arXiv:2112.03126}, 2021.

\bibitem[Birkl et~al.(2023)Birkl, Wofk, and M{\"u}ller]{birkl2023midas}
Reiner Birkl, Diana Wofk, and Matthias M{\"u}ller.
\newblock Midas v3.1 -- a model zoo for robust monocular relative depth estimation.
\newblock \emph{arXiv preprint arXiv:2307.14460}, 2023.

\bibitem[Blanz and Vetter(2023)]{blanz2023morphable}
Volker Blanz and Thomas Vetter.
\newblock A morphable model for the synthesis of 3d faces.
\newblock In \emph{Seminal Graphics Papers: Pushing the Boundaries, Volume 2}, pages 157--164. 2023.

\bibitem[Brooks et~al.(2023)Brooks, Holynski, and Efros]{brooks2022instructpix2pix}
Tim Brooks, Aleksander Holynski, and Alexei~A. Efros.
\newblock Instructpix2pix: Learning to follow image editing instructions.
\newblock In \emph{CVPR}, 2023.

\bibitem[Deitke et~al.(2023)Deitke, Liu, Wallingford, Ngo, Michel, Kusupati, Fan, Laforte, Voleti, Gadre, et~al.]{deitke2023objaverse}
Matt Deitke, Ruoshi Liu, Matthew Wallingford, Huong Ngo, Oscar Michel, Aditya Kusupati, Alan Fan, Christian Laforte, Vikram Voleti, Samir~Yitzhak Gadre, et~al.
\newblock Objaverse-xl: A universe of 10m+ 3d objects.
\newblock \emph{arXiv preprint arXiv:2307.05663}, 2023.

\bibitem[Dhariwal and Nichol(2021)]{dhariwal2021diffusion}
Prafulla Dhariwal and Alexander Nichol.
\newblock Diffusion models beat gans on image synthesis.
\newblock \emph{NeurIPS}, 2021.

\bibitem[Downs et~al.(2022)Downs, Francis, Koenig, Kinman, Hickman, Reymann, McHugh, and Vanhoucke]{downs2022google}
Laura Downs, Anthony Francis, Nate Koenig, Brandon Kinman, Ryan Hickman, Krista Reymann, Thomas~B. McHugh, and Vincent Vanhoucke.
\newblock Google scanned objects: A high-quality dataset of {3D} scanned household items.
\newblock In \emph{ICRA}, 2022.

\bibitem[Ehsani et~al.(2018)Ehsani, Mottaghi, and Farhadi]{ehsani2018segan}
Kiana Ehsani, Roozbeh Mottaghi, and Ali Farhadi.
\newblock Segan: Segmenting and generating the invisible.
\newblock In \emph{CVPR}, 2018.

\bibitem[Gal et~al.(2022)Gal, Alaluf, Atzmon, Patashnik, Bermano, Chechik, and Cohen-Or]{gal2022image}
Rinon Gal, Yuval Alaluf, Yuval Atzmon, Or Patashnik, Amit~H Bermano, Gal Chechik, and Daniel Cohen-Or.
\newblock An image is worth one word: Personalizing text-to-image generation using textual inversion.
\newblock \emph{arXiv preprint arXiv:2208.01618}, 2022.

\bibitem[Goodfellow et~al.(2014)Goodfellow, Pouget-Abadie, Mirza, Xu, Warde-Farley, Ozair, Courville, and Bengio]{goodfellow2014generative}
Ian Goodfellow, Jean Pouget-Abadie, Mehdi Mirza, Bing Xu, David Warde-Farley, Sherjil Ozair, Aaron Courville, and Yoshua Bengio.
\newblock Generative adversarial nets.
\newblock \emph{NeurIPS}, 2014.

\bibitem[Ho and Salimans(2022)]{ho2022classifier}
Jonathan Ho and Tim Salimans.
\newblock Classifier-free diffusion guidance.
\newblock \emph{arXiv preprint arXiv:2207.12598}, 2022.

\bibitem[Ho et~al.(2020)Ho, Jain, and Abbeel]{ho2020denoising}
Jonathan Ho, Ajay Jain, and Pieter Abbeel.
\newblock Denoising diffusion probabilistic models.
\newblock \emph{NeurIPS}, 33, 2020.

\bibitem[Hsieh et~al.(2023)Hsieh, Khurana, Dave, and Ramanan]{hsieh2023tracking}
Cheng-Yen Hsieh, Tarasha Khurana, Achal Dave, and Deva Ramanan.
\newblock Tracking any object amodally, 2023.

\bibitem[Hu et~al.(2019)Hu, Chen, Hui, Huang, and Schwing]{HuCVPR2019}
Y.-T. Hu, H.-S. Chen, K. Hui, J.-B. Huang, and A.~G. Schwing.
\newblock {SAIL-VOS: Semantic Amodal Instance Level Video Object Segmentation -- A Synthetic Dataset and Baselines}.
\newblock In \emph{Proc. CVPR}, 2019.

\bibitem[Kar et~al.(2015)Kar, Tulsiani, Carreira, and Malik]{kar2015amodal}
Abhishek Kar, Shubham Tulsiani, Joao Carreira, and Jitendra Malik.
\newblock Amodal completion and size constancy in natural scenes.
\newblock In \emph{ICCV}, 2015.

\bibitem[Ke et~al.(2021)Ke, Tai, and Tang]{ke2021deep}
Lei Ke, Yu-Wing Tai, and Chi-Keung Tang.
\newblock Deep occlusion-aware instance segmentation with overlapping bilayers.
\newblock In \emph{CVPR}, 2021.

\bibitem[Kingma and Welling(2013)]{kingma2013auto}
Diederik~P Kingma and Max Welling.
\newblock Auto-encoding variational bayes.
\newblock \emph{arXiv preprint arXiv:1312.6114}, 2013.

\bibitem[Kirillov et~al.(2023)Kirillov, Mintun, Ravi, Mao, Rolland, Gustafson, Xiao, Whitehead, Berg, Lo, Dollár, and Girshick]{kirillov2023segment}
Alexander Kirillov, Eric Mintun, Nikhila Ravi, Hanzi Mao, Chloe Rolland, Laura Gustafson, Tete Xiao, Spencer Whitehead, Alexander~C. Berg, Wan-Yen Lo, Piotr Dollár, and Ross Girshick.
\newblock Segment anything.
\newblock In \emph{ICCV}, 2023.

\bibitem[Ling et~al.(2020)Ling, Acuna, Kreis, Kim, and Fidler]{ling2020variational}
Huan Ling, David Acuna, Karsten Kreis, Seung~Wook Kim, and Sanja Fidler.
\newblock Variational amodal object completion.
\newblock \emph{NeurIPS}, 2020.

\bibitem[Liu and Vondrick(2023)]{liu2023humans}
Ruoshi Liu and Carl Vondrick.
\newblock Humans as light bulbs: 3d human reconstruction from thermal reflection.
\newblock In \emph{CVPR}, 2023.

\bibitem[Liu et~al.(2022)Liu, Menon, Mao, Park, Stent, and Vondrick]{liu2022shadows}
Ruoshi Liu, Sachit Menon, Chengzhi Mao, Dennis Park, Simon Stent, and Carl Vondrick.
\newblock Shadows shed light on 3d objects.
\newblock \emph{arXiv preprint arXiv:2206.08990}, 2022.

\bibitem[Liu et~al.(2023{\natexlab{a}})Liu, Mao, Tendulkar, Wang, and Vondrick]{liu2023landscape}
Ruoshi Liu, Chengzhi Mao, Purva Tendulkar, Hao Wang, and Carl Vondrick.
\newblock Landscape learning for neural network inversion.
\newblock In \emph{ICCV}, 2023{\natexlab{a}}.

\bibitem[Liu et~al.(2023{\natexlab{b}})Liu, Wu, Van~Hoorick, Tokmakov, Zakharov, and Vondrick]{liu2023zero}
Ruoshi Liu, Rundi Wu, Basile Van~Hoorick, Pavel Tokmakov, Sergey Zakharov, and Carl Vondrick.
\newblock Zero-1-to-3: Zero-shot one image to 3d object.
\newblock In \emph{ICCV}, 2023{\natexlab{b}}.

\bibitem[Liu et~al.(2023{\natexlab{c}})Liu, Lin, Zeng, Long, Liu, Komura, and Wang]{liu2023syncdreamer}
Yuan Liu, Cheng Lin, Zijiao Zeng, Xiaoxiao Long, Lingjie Liu, Taku Komura, and Wenping Wang.
\newblock Syncdreamer: Learning to generate multiview-consistent images from a single-view image.
\newblock \emph{arXiv preprint arXiv:2309.03453}, 2023{\natexlab{c}}.

\bibitem[Ma et~al.(2022)Ma, Wang, Yuille, and Kortylewski]{ma2022robust}
Wufei Ma, Angtian Wang, Alan Yuille, and Adam Kortylewski.
\newblock Robust category-level {6D} pose estimation with coarse-to-fine rendering of neural features.
\newblock In \emph{ECCV}, 2022.

\bibitem[Martin et~al.(2001)Martin, Fowlkes, Tal, and Malik]{MartinFTM01}
D. Martin, C. Fowlkes, D. Tal, and J. Malik.
\newblock A database of human segmented natural images and its application to evaluating segmentation algorithms and measuring ecological statistics.
\newblock In \emph{ICCV}, 2001.

\bibitem[Mildenhall et~al.(2020)Mildenhall, Srinivasan, Tancik, Barron, Ramamoorthi, and Ng]{mildenhall2020nerf}
Ben Mildenhall, Pratul~P. Srinivasan, Matthew Tancik, Jonathan~T. Barron, Ravi Ramamoorthi, and Ren Ng.
\newblock Nerf: Representing scenes as neural radiance fields for view synthesis.
\newblock In \emph{ECCV}, 2020.

\bibitem[Piaget(2013)]{piaget2013construction}
Jean Piaget.
\newblock \emph{The construction of reality in the child}.
\newblock Routledge, 2013.

\bibitem[Podell et~al.(2023)Podell, English, Lacey, Blattmann, Dockhorn, Müller, Penna, and Rombach]{podell2023sdxl}
Dustin Podell, Zion English, Kyle Lacey, Andreas Blattmann, Tim Dockhorn, Jonas Müller, Joe Penna, and Robin Rombach.
\newblock Sdxl: Improving latent diffusion models for high-resolution image synthesis, 2023.

\bibitem[Poole et~al.(2022)Poole, Jain, Barron, and Mildenhall]{poole2022dreamfusion}
Ben Poole, Ajay Jain, Jonathan~T Barron, and Ben Mildenhall.
\newblock Dreamfusion: Text-to-3d using 2d diffusion.
\newblock \emph{arXiv preprint arXiv:2209.14988}, 2022.

\bibitem[Qi et~al.(2019)Qi, Jiang, Liu, Shen, and Jia]{qi2019amodal}
Lu Qi, Li Jiang, Shu Liu, Xiaoyong Shen, and Jiaya Jia.
\newblock Amodal instance segmentation with {KINS} dataset.
\newblock In \emph{CVPR}, 2019.

\bibitem[Radford et~al.(2021)Radford, Kim, Hallacy, Ramesh, Goh, Agarwal, Sastry, Askell, Mishkin, Clark, Krueger, and Sutskever]{radford2021learning}
Alec Radford, Jong~Wook Kim, Chris Hallacy, Aditya Ramesh, Gabriel Goh, Sandhini Agarwal, Girish Sastry, Amanda Askell, Pamela Mishkin, Jack Clark, Gretchen Krueger, and Ilya Sutskever.
\newblock Learning transferable visual models from natural language supervision, 2021.

\bibitem[Reddy et~al.(2022)Reddy, Tamburo, and Narasimhan]{reddy2022walt}
N~Dinesh Reddy, Robert Tamburo, and Srinivasa~G Narasimhan.
\newblock Walt: Watch and learn 2d amodal representation from time-lapse imagery.
\newblock In \emph{CVPR}, 2022.

\bibitem[Rombach et~al.(2022)Rombach, Blattmann, Lorenz, Esser, and Ommer]{rombach2022high}
Robin Rombach, Andreas Blattmann, Dominik Lorenz, Patrick Esser, and Bj{\"o}rn Ommer.
\newblock High-resolution image synthesis with latent diffusion models.
\newblock In \emph{CVPR}, 2022.

\bibitem[Ruiz et~al.(2023)Ruiz, Li, Jampani, Pritch, Rubinstein, and Aberman]{ruiz2023dreambooth}
Nataniel Ruiz, Yuanzhen Li, Varun Jampani, Yael Pritch, Michael Rubinstein, and Kfir Aberman.
\newblock Dreambooth: Fine tuning text-to-image diffusion models for subject-driven generation.
\newblock In \emph{CVPR}, 2023.

\bibitem[Sargent et~al.(2023)Sargent, Li, Shah, Herrmann, Yu, Zhang, Chan, Lagun, Fei-Fei, Sun, and Wu]{zeronvs}
Kyle Sargent, Zizhang Li, Tanmay Shah, Charles Herrmann, Hong-Xing Yu, Yunzhi Zhang, Eric~Ryan Chan, Dmitry Lagun, Li Fei-Fei, Deqing Sun, and Jiajun Wu.
\newblock {ZeroNVS}: Zero-shot 360-degree view synthesis from a single real image.
\newblock \emph{arXiv preprint arXiv:2310.17994}, 2023.

\bibitem[Schuhmann et~al.(2022)Schuhmann, Beaumont, Vencu, Gordon, Wightman, Cherti, Coombes, Katta, Mullis, Wortsman, et~al.]{schuhmann2022laion}
Christoph Schuhmann, Romain Beaumont, Richard Vencu, Cade Gordon, Ross Wightman, Mehdi Cherti, Theo Coombes, Aarush Katta, Clayton Mullis, Mitchell Wortsman, et~al.
\newblock Laion-{5B}: An open large-scale dataset for training next generation image-text models.
\newblock \emph{NeurIPS}, 2022.

\bibitem[Shtedritski et~al.(2023)Shtedritski, Rupprecht, and Vedaldi]{shtedritski2023does}
Aleksandar Shtedritski, Christian Rupprecht, and Andrea Vedaldi.
\newblock What does clip know about a red circle? visual prompt engineering for vlms.
\newblock In \emph{ICCV}, 2023.

\bibitem[Song et~al.(2020)Song, Meng, and Ermon]{song2020denoising}
Jiaming Song, Chenlin Meng, and Stefano Ermon.
\newblock Denoising diffusion implicit models.
\newblock \emph{arXiv preprint arXiv:2010.02502}, 2020.

\bibitem[Tu et~al.(2005)Tu, Chen, Yuille, and Zhu]{tu2005image}
Zhuowen Tu, Xiangrong Chen, Alan~L Yuille, and Song-Chun Zhu.
\newblock Image parsing: Unifying segmentation, detection, and recognition.
\newblock \emph{International Journal of computer vision}, 63:\penalty0 113--140, 2005.

\bibitem[Wang et~al.(2021)Wang, Liu, Liu, Theobalt, Komura, and Wang]{wang2021neus}
Peng Wang, Lingjie Liu, Yuan Liu, Christian Theobalt, Taku Komura, and Wenping Wang.
\newblock Neus: Learning neural implicit surfaces by volume rendering for multi-view reconstruction.
\newblock \emph{arXiv preprint arXiv:2106.10689}, 2021.

\bibitem[Wang et~al.(2004)Wang, Bovik, Sheikh, and Simoncelli]{wang2004image}
Zhou Wang, Alan~C Bovik, Hamid~R Sheikh, and Eero~P Simoncelli.
\newblock Image quality assessment: from error visibility to structural similarity.
\newblock \emph{IEEE Transactions on Image Processing}, 13\penalty0 (4):\penalty0 600--612, 2004.

\bibitem[Wu et~al.(2023)Wu, Liu, Vondrick, and Zheng]{wu2023sin3dm}
Rundi Wu, Ruoshi Liu, Carl Vondrick, and Changxi Zheng.
\newblock Sin3dm: Learning a diffusion model from a single 3d textured shape.
\newblock \emph{arXiv preprint arXiv:2305.15399}, 2023.

\bibitem[Xu et~al.(2023)Xu, Liu, Vahdat, Byeon, Wang, and De~Mello]{xu2023odise}
Jiarui Xu, Sifei Liu, Arash Vahdat, Wonmin Byeon, Xiaolong Wang, and Shalini De~Mello.
\newblock {Open-Vocabulary Panoptic Segmentation with Text-to-Image Diffusion Models}.
\newblock \emph{arXiv preprint arXiv:2303.04803}, 2023.

\bibitem[Yuille and Kersten(2006)]{yuille2006vision}
Alan Yuille and Daniel Kersten.
\newblock Vision as bayesian inference: analysis by synthesis?
\newblock \emph{Trends in cognitive sciences}, 10\penalty0 (7):\penalty0 301--308, 2006.

\bibitem[Zhan et~al.(2022)Zhan, Xie, and Zisserman]{zhan2022triocc}
Guanqi Zhan, Weidi Xie, and Andrew Zisserman.
\newblock A tri-layer plugin to improve occluded detection.
\newblock \emph{BMVC}, 2022.

\bibitem[Zhan et~al.(2020)Zhan, Pan, Dai, Liu, Lin, and Loy]{zhan2020self}
Xiaohang Zhan, Xingang Pan, Bo Dai, Ziwei Liu, Dahua Lin, and Chen~Change Loy.
\newblock Self-supervised scene de-occlusion.
\newblock In \emph{CVPR}, 2020.

\bibitem[Zhang et~al.(2018)Zhang, Isola, Efros, Shechtman, and Wang]{zhang2018perceptual}
Richard Zhang, Phillip Isola, Alexei~A Efros, Eli Shechtman, and Oliver Wang.
\newblock The unreasonable effectiveness of deep features as a perceptual metric.
\newblock In \emph{CVPR}, 2018.

\bibitem[Zhang et~al.(2023)Zhang, Ji, Wang, Mei, Kortylewski, and Yuille]{zhang20233d}
Yi Zhang, Pengliang Ji, Angtian Wang, Jieru Mei, Adam Kortylewski, and Alan Yuille.
\newblock {3D-Aware} neural body fitting for occlusion robust 3d human pose estimation.
\newblock In \emph{ICCV}, 2023.

\bibitem[Zhu et~al.(2020)Zhu, Shen, Zhao, and Zhou]{zhu2020domain}
Jiapeng Zhu, Yujun Shen, Deli Zhao, and Bolei Zhou.
\newblock In-domain {GAN} inversion for real image editing.
\newblock In \emph{ECCV}, 2020.

\bibitem[Zhu et~al.(2017)Zhu, Tian, Metaxas, and Doll{\'a}r]{zhu2016semantic}
Yan Zhu, Yuandong Tian, Dimitris Metaxas, and Piotr Doll{\'a}r.
\newblock Semantic amodal segmentation.
\newblock In \emph{CVPR}, 2017.

\end{thebibliography}
}


\end{document}